\documentclass[sigconf]{acmart}

\usepackage{times}
\usepackage{epsfig}
\usepackage{graphicx}
\usepackage{amsmath}
\usepackage[utf8]{inputenc} 
\usepackage[T1]{fontenc}    
\usepackage{url}            
\usepackage{booktabs}       
\usepackage{amsfonts}       
\usepackage{nicefrac}       
\usepackage{microtype}      
\usepackage{xcolor}         
\usepackage{array}          
\usepackage{multirow}       

\AtBeginDocument{%
  }
\setcopyright{none}
\renewcommand\footnotetextcopyrightpermission[1]{}

\begin{document}

\title{Improving the Accuracy of Beauty Product Recommendations by Assessing Face Illumination Quality}


\author{Parnian Afshar$^*$, Jenny Yeon$^*$, Andriy Levitskyy, Rahul Suresh, and Amin Banitalebi-Dehkordi}

\email{{pafshar, jenyeon, levitsan, surerahu, aminbt}@amazon.com}

\affiliation{%
  \institution{Amazon}
  \city{Vancouver}
  \country{Canada}}

\renewcommand{\shortauthors}{Afshar et al.}

\begin{abstract}
We focus on addressing the challenges in responsible beauty product recommendation, particularly when it involves comparing the product's color with a person's skin tone, such as for foundation and concealer products. To make accurate recommendations, it is crucial to infer both the product attributes and the product specific facial features such as skin conditions or tone. However, while many product photos are taken under good light conditions, face photos are taken from a wide range of conditions. The features extracted using the photos from ill-illuminated environment can be highly misleading or even be incompatible to be compared with the product attributes. Hence bad illumination condition can severely degrade quality of the recommendation.


We introduce a machine learning framework for \textbf{illumination assessment} which classifies images into having either good or bad illumination condition. We then build an automatic user guidance tool which informs a user holding their camera if their illumination condition is good or bad. This way, the user is provided with rapid feedback and can interactively control how the photo is taken for their recommendation. Only a few studies are dedicated to this problem, mostly due to the lack of dataset that is large, labeled, and diverse both in terms of skin tones and light patterns. Lack of such dataset leads to neglecting skin tone diversity. Therefore, We begin by constructing a diverse synthetic dataset that simulates various skin tones and light patterns in addition to an existing facial image dataset. Next, we train a Convolutional Neural Network (CNN) for illumination assessment that outperforms the existing solutions using the synthetic dataset. Finally, we analyze how the our work improves the \textbf{shade recommendation} for various foundation products.

\end{abstract}

\maketitle
\def\thefootnote{*}\footnotetext{Equal contribution.}
\pagestyle{plain}

\section{Introduction}
\begin{figure}
    \centering
    \includegraphics[scale=0.1]{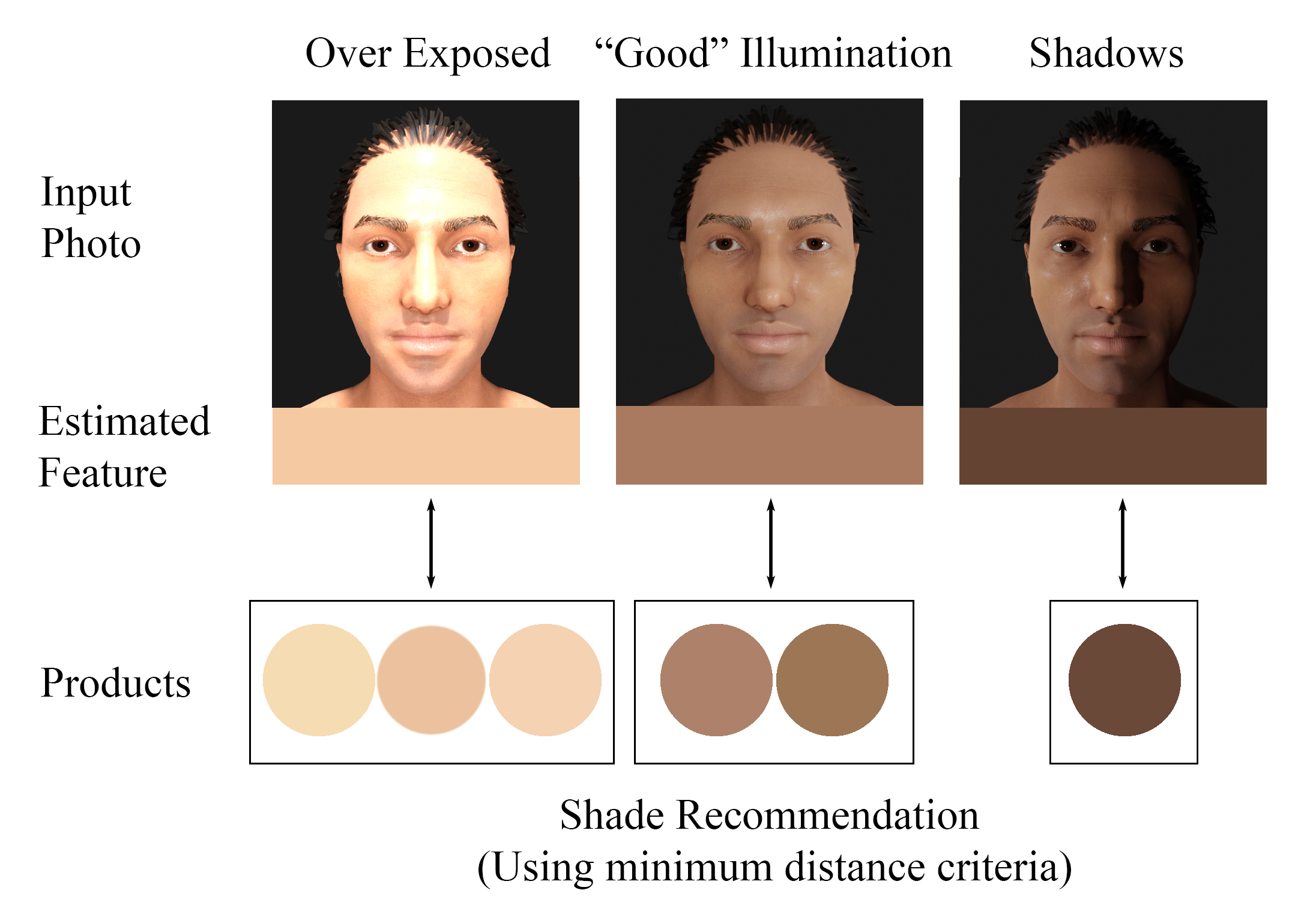}
    \vspace{-12pt}
    \caption{\small{Effect of using photos under different illumination conditions on shade recommendation for foundation and concealer products. Only well-illuminated face photo (middle) results in recommending medium range shades as intended. The same user is given conflicting recommendations in the other two cases. Synthetic face images are used because of privacy reasons and having more control over illumination/ skin tone variations.}}
    \label{fig:rec_big_picture}
\vspace{-20pt}
\end{figure}

Many beauty product recommendation systems require information on both the products and customers' preferences~\cite{Nguyen}. Traditionally, customers' preferences are collected through series of questions and answers. However, many customers do not know how to answer such specialized beauty-related questions, and the recommendation produced may not feel worth the time they spent ~\cite{Liu}. On the other hand, predictive models to infer information about the customers may not lead to accurate results. An alternative route to reduce the barrier for beauty product recommendation is to use camera images as input and provide skin care or makeup recommendations based on the facial features estimated from the images.
However, it is challenging to extract facial features from photos unless they are taken under a good illumination~\cite{b1} condition. The user may not be paying attention to the illumination condition while taking photos, but the photos taken under poor illumination conditions highly impact how face features such as skin conditions, makeup, tone, etc. are inferred. See Figure \ref{fig:rec_big_picture}. The products recommended based on conflicting or incorrect feature values resulted from a set of poorly illuminated face photos may even provide a recommendation that is worse than recommending random products. 

While tracking the face illumination condition is highly important for improving recommendation, only a few studies have been conducted to address this problem resulting in a few automatic models available for this purpose~\cite{b1}. Our approach is to apply an automatic model to directly inform the user holding a camera to take picture if the light condition is good or bad, which allows the user to be selective about what photo to be used for the recommendation. Specifically, the tool displays the outcome of the illumination assessment model (either good or bad) to the user holding their camera, which then encourages them to interactively change the camera position or environment until they see the light condition being good. One alternative approach is to take any image from the user and attempt to correctly infer features \cite{Malherbe_2022_ACCV}. However, some studies do not consider images taken under poor illumination conditions or require images that meet special requirements, such as holding a color calibration card. It is also unclear how much feature inference varies if differently illuminated photos of the same person are used by the method. While these existing solutions may reduce the time needed for the user to interact with their camera, our approach gives the user a direct control over how their photo would be illuminated. Moreover, showing the output of the model and allowing interactivity may help increase trustworthiness \cite{Ahn2021WillWT, Honeycutt_Nourani_Ragan_2020}. 

\subsection{Related Work}
One of the studies on face illumination quality assessment is conducted by Sellahewa and Jassim~\cite{b2}. Their method, however, is limited by requiring a reference image which reduces the applicability of their approach. Another method by Truong \textit{et al.}~\cite{b3} approaches illumination assessment by averaging over image partitions. This technique is too simple to account for diverse illumination scenarios. Similar pixel averaging is also used in a recent study by Hernandez-Ortega \textit{et al.}~\cite{Hernandez-Ortega}. In another study by Terhorst \textit{et al.}~\cite{Terhorst}, pixel-level face quality is considered rather than image-level, which is considered in our study.

Many other prior studies have focused on face quality assessment in general, i.e., considering different factors such as head pose, noise, sharpness, and illumination. One of such studies is performed by Chen \textit{et al.}~\cite{b4}, to select images of high quality for the downstream task of face recognition. This group of studies however are not able to provide an explicit signal about the bad illumination condition, which is the main goal of this study.



One of the challenges that has prevented the development of an automatic illumination assessment model is the lack of a large dataset of faces under various lighting conditions. Existing face datasets are mostly under neutral illumination, or captured in-the-wild with uncontrolled illumination. The ones that include images of various lighting conditions are either not labeled or limited to a small number of subjects. To fill this gap, Zhang \textit{et al.}~\cite{b1} constructed a large-scale dataset of face images with synthetic illumination patterns, labeled based on their quality. To construct the dataset, one subject took photos at various lighting conditions. These patterns were then transferred to existing face datasets such as YaleB~\cite{b5}, PIE~\cite{b6}, and FERET~\cite{b7}. To transfer light patterns, an illumination transfer algorithm through edge-preserving filters proposed in~\cite{b8} was used. This technique however works under the assumption of similar face shape between source and target, it requires frontal faces, and fails under intense lighting conditions. Another drawback of the study by Zhang \textit{et al.}~\cite{b1} is that their illumination assessment model fails when fed with dark skin tone faces and classifies most of them as bad lighting. As discussed in a study by Babnik and Struc, face quality assessment models highly favor light-colored faces~\cite{Babnik}.
 Disentangling skin tone and illumination is still an open problem~\cite{Feng}, and machine learning models are not able to assess illumination quality unless they are either trained on an inclusive and representative dataset, or fed with background information~\cite{Feng}. Asking the users to include background in their photos, however, results in low resolution faces, for most of the mobile devices, reducing the performance of the consequent face processing applications.
 \vspace{-8pt}
\subsection{Contributions}
Achieving a representative dataset is costly and time-consuming. For this reason, in this work, we propose to first increase the diversity of a face dataset by simulating light to dark skin tones. Simulating skin tones has the benefit of increasing the diversity without  collecting such data. consequently, 200 illumination patterns, including good and bad ones, are defined and transferred to the dataset, using the Deep Single-Image Portrait Relighting (DPR) model~\cite{b9}. This model is able to transfer a target illumination to a source portrait image. Unlike Reference~\cite{b8}, DPR is not restricted to frontal and same shape source and reference faces, and it can handle harsh lighting. Its limitation, however, is that light and skin tone are not perfectly disentangled and some skin color is transferred along illumination. Therefore, for each skin tone, we transfer illumination defined on the same tone.  The constructed dataset is then used to train a MobileNet-v2~\cite{b10} model to classify faces as good/bad illuminated. 

The proposed illumination assessment model is used in improving beauty product recommendation based on user's face photo. We illustrate how the retrieved product shade is significantly different than the shade we should retrieve based on the real face features such as skin tone, under proper illumination.

The rest of this paper is organized as follows: our proposed framework, including illumination assessment and beauty product recommendation, is described in Section~\ref{sec:framework}, followed by the experiments in Section~\ref{sec:experiment}. 
Finally, Section~\ref{sec:conclusion} concludes the study. 
\vspace{-3pt}
\section{The Proposed Framework}\label{sec:framework}
\begin{figure*}
\centerline{\includegraphics[scale=0.25]{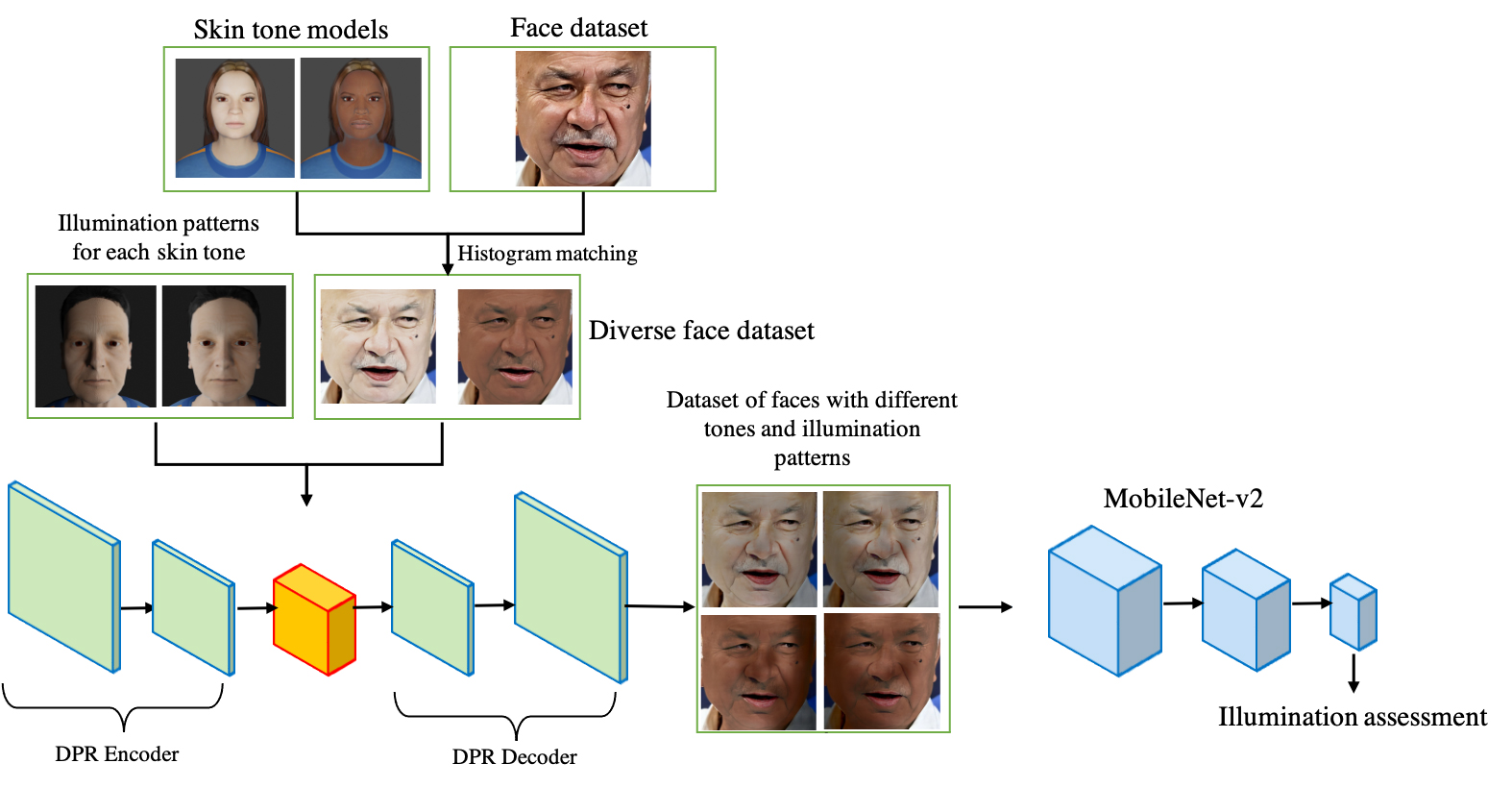}}
\vspace{-12pt}
\caption{\small{Proposed framework. Light and dark skin tones are transferred to detected faces, followed by applying different illumination patterns. Original face photo from~\cite{vgg}, used for illustration purposes only.}}
\label{fig:framework}
\end{figure*}
Our proposed framework, explained in the following sections, includes face illumination assessment, and using the properly illuminated face, for beauty product recommendation.
\subsection{Face Illumination Assessment}

Our face illumination assessment workflow is illustrated in Fig.~\ref{fig:framework}. At the first step, a 3D modelling software is used to create models of different skin tone and illumination patterns, which are labeled based on their quality for the downstream tasks such as extracting face features. The skin tones, light patterns and their associated labels are consequently transferred to an in-house dataset of 1000 frontal faces, mostly under neutral illumination. Finally a MobileNet-v2 is trained on the constructed dataset to predict two classes of good and bad illumination. These steps are described in more detail in the following subsections.

\subsubsection{Generating synthetic examples of skin tones and illumination patterns}

\begin{figure}
\centerline{\includegraphics[scale=0.12]{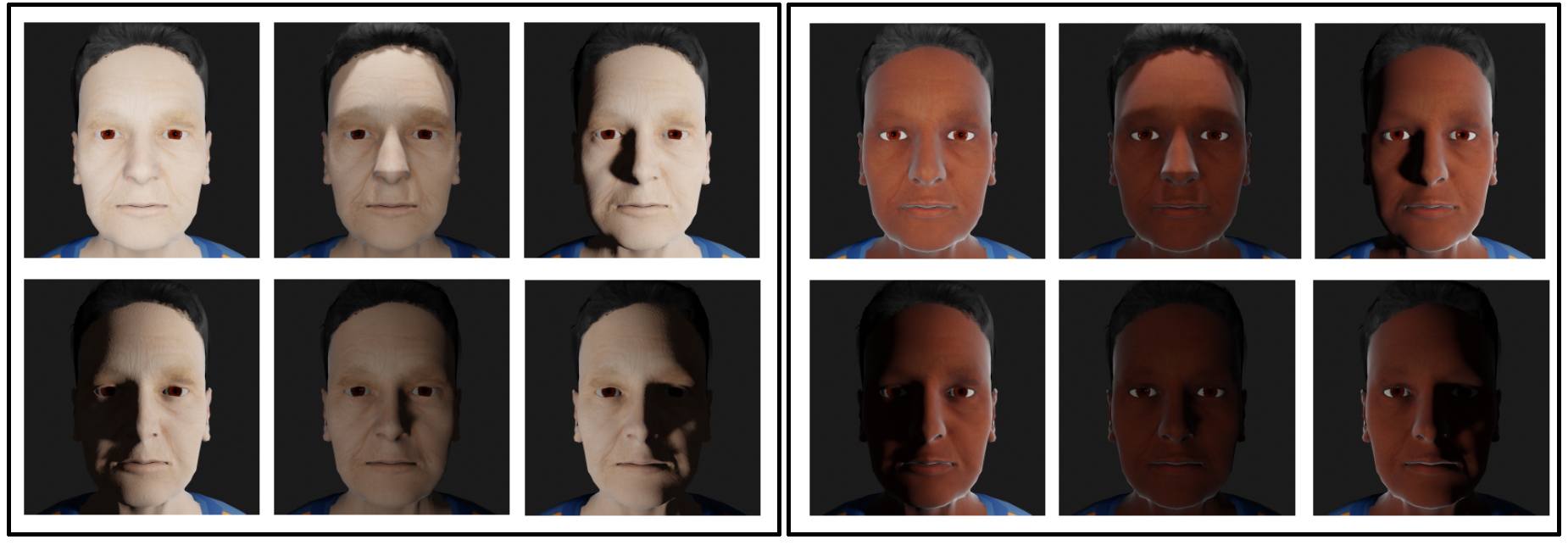}}
\vspace{-12pt}
\caption{\small{Examples of light and dark skin tone images with different illumination patterns.}}
\label{fig:light-dark}
\vspace{-8pt}
\end{figure}


We used a 3D modelling software to generate models of light and dark skin tones as well as 200 illumination patterns, labeled based on their quality. The skin tones and patterns are all transferred to the in-house dataset of 1000 faces. The labels for the constructed dataset directly come from the pattern labels, which means that the only manual labeling required for this study was to label the 200 light patterns generated by a 3D modelling software. Fig.~\ref{fig:light-dark} illustrates some examples for both light and dark skin tones.

The light patterns were generated by moving two spot light sources among the grid of coordinate values relative to a model of human. The grid covers situations where the light sources illuminate face from the side and/or are located above, below and in front of the person. The different skin-tones were obtained by linearly combining textures of light and dark-skinned models.

It is also worth noting that the 3D modelling software used in this study generates mostly  unrealistic cartoon images as shown in Fig.~\ref{fig:light-dark}. Therefore, these images cannot be directly used for training, and the consequent color matching and light transfer to a real face dataset is necessary. There are 3D modelling software available that can generate more realistic face images. Those software, however, require more rendering time and computational resources, are limited to a few hundreds human models, and acquiring their high quality models is expensive. 

\subsubsection{Color matching to simulate skin tones}

To transfer skin tone from a source to a target image, we first segmented the skin area of the face, using the BiSeNet model introduced in Reference~\cite{b12}. BiSeNet is a real time semantic segmentation architecture with two branches, the first of which captures low-level details, and the second captures high-level context. Segmenting the face skin area is followed by a histogram matching technique~\cite{b13,b14}. To perform histogram matching, first the cumulative color histogram of the source and reference images, denoted by $C_s$ and $C_r$ are calculated, where source and reference refer to the face image and the skin tone model, respectively. Having the cumulative histogram functions, the output image $\textbf{I}_o$ for pixel $p$ is derived from the input image $\textbf{I}$, as:

\begin{equation}
\textbf{I}_o(p)=v_r \left(C_r^{-1} \left(C_s \left( \frac{\textbf{I}(p)- \text{min}(\textbf{I}) + 1}{V}    \right) \right) \right),
\end{equation}

\noindent where $C_r^{-1}$ acts as a reverse lookup on the histogram, $V$ is the histogram bin width, and function $v$ is defined as:

\begin{equation}
v (i) = \text{min}(\textbf{I}) + (i-1)V.
\end{equation}

\subsubsection{DPR model to transfer illumination patterns}

DPR is an hour-glass~\cite{b15} architecture that takes a Spherical Harmonic (SH) lighting as the target illumination, and a portrait image as the source image. Consequently the target lighting is applied to the source image to generate the output. Besides the output image, DPR also generates the SH lighting parameters of the input portrait. Since in this study we did not have the SH lighting of the reference illumination patterns, we first fed all the reference images to DPR to get the associated SH lighting. These parameters are then used as target illumination to be applied to all the source images.

One limitation of the DPR model is that it cannot completely disentangle lighting and skin tone, as in general what we see from a face image is influenced by both the inherent skin tone as well as environment illumination and it is difficult to disentangle the two. In other words, using a light tone reference for a dark tone source image results in a non-realistic face. To tackle this, for light skin tone faces, illumination patterns are transferred from the light skin model, and for dark skin faces, patterns are transformed from the dark skin model. 
\subsubsection{MediaPipe face detection and MobileNet-v2}

In this study, we have used MediaPipe~\cite{b16} solution for face detection, in order to omit irrelevant background information. MediaPipe face detection is built upon BlazeFace~\cite{b17} which is a light-weight face detection architecture. 

Finally, the constructed dataset is fed to a MobileNet-v2 model to classify faces as good or bad illuminated. This architecture is selected for its efficiency to run on edge devices. To handle class imbalance caused by more bad illumination patterns than good ones (because in general good illumination is just one pattern i.e. no shadow or over-exposure on the face), more weight is given to error on good illuminated faces, in the cross-entropy loss function.


\subsection{Using Face Photos for Beauty Product Recommendation}

\begin{figure*}
\centerline{\includegraphics[scale=0.25]{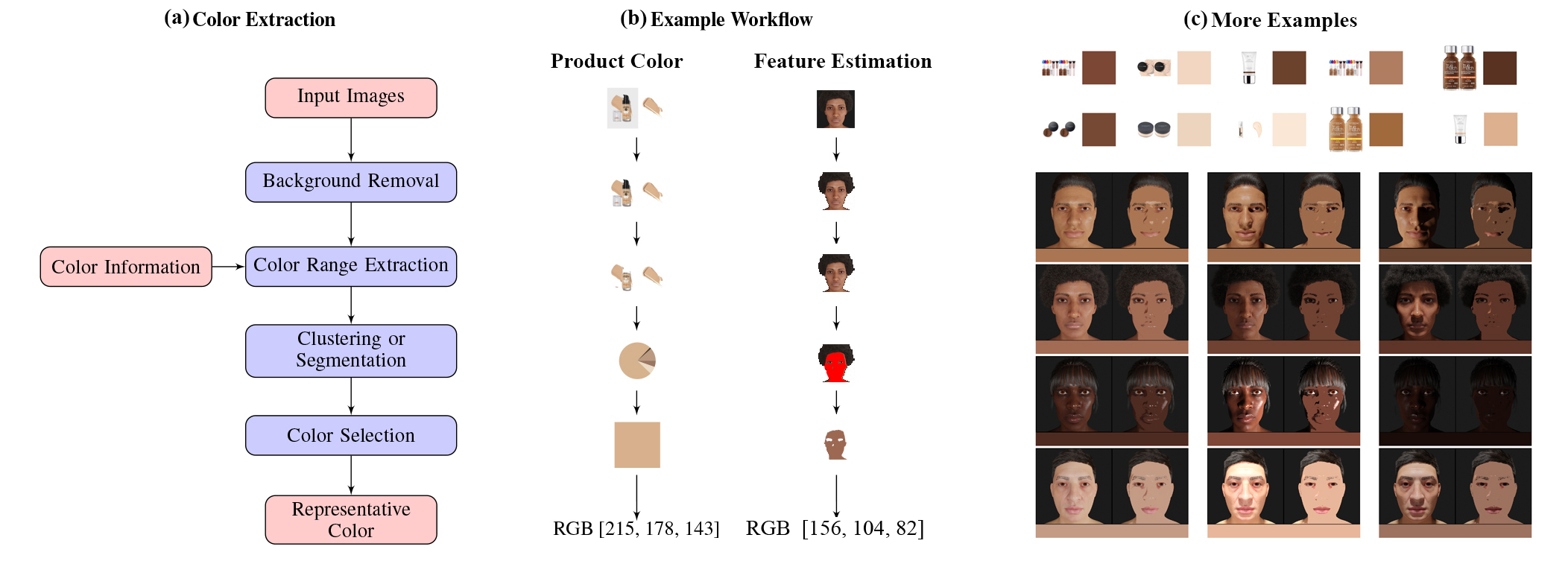}}
\vspace{-12pt}
\caption{\small{Color Extraction Method. (a) Overview of how to extract representative color from images. 
(b) Product color is extracted by taking dominant brown color shown in the image after the color range extraction. The estimated skin tone is computed as the average value of the pixels in the face and the front neck area. (c) Examples of the colors extracted from the product images (the first two rows) and the synthetic face photos. Each row of the face photos is the same person but under different illumination patterns. }}
\label{fig:foundation}
\vspace{-8pt}
\end{figure*}
While some beauty product recommendation does not require knowing facial features such as skin tone, it is difficult to recommend base makeup products such as foundation and concealers without knowing such features. Foundation products in particular have an attribute called \textit{shade} which denotes the color of the product. A foundation product is offered in multiple shades, and \textit{shade recommendation} comprises of selecting the shade that best matches a customer's need. See Figure \ref{fig:rec_big_picture}. For example, some customers may prefer a shade that is closest to their skin tone, while others prefer a shade that is either lighter/darker or warmer/cooler than their skin tone. Regardless of different preferences, the color difference between each shade and the customer's skin tone needs to be assessed in order to proceed with the shade recommendation. Many customers feel overwhelmed in selecting the shade that matches their preference, especially if they have to shop online. In this section, we will first describe how we assign an RGB value for each foundation shade, and how we compute \textit{estimated skin tone} from face photos. Lastly, we will introduce how we compare the shade color with the estimated skin tone. In Section \ref{sec:usecase_exp}, we will show how our illumination assessment improves the quality of the shade recommendation. 
\vspace{-10pt}
\subsubsection{Product Color and Facial Feature Estimation}
Historically, foundation shade range has not been adequately reflecting skin tone diversity and only in recent years, this issue has been actively corrected by the industry \cite{pudding, shades, luxury}. This also means that the shade distribution tends to be heavy on the lighter color spectrum, and a color inference model trained on unmitigated training dataset can be prone to bias. Therefore, we applied the product color extraction framework shown in Fig.~\ref{fig:foundation} using an unsupervised approach for each component. Specifically, we applied threshold based background removal, and extract colors in certain fixed range, and applied K-means to extract the most prominent brown color shown in the foundation product images. Using this approach, we obtained about 2000 shades of foundation products. Any error cases were examined and removed from the dataset.
A similar color extraction framework was adapted for face images in estimating skin tones. That is, we remove the background and extract the color using the relevant segment of the photo, as shown in Fig.~\ref{fig:foundation}. We call this resulting color value estimated skin tone. It is important to note that we do not claim that the estimated skin tone is the person's actual skin tone. We only infer the skin tone as they are represented as RGB values in the photos in order to compare it with the product colors to produce a meaningful shade recommendation. Poor illumination conditions such as partial shadows can affect the value of estimated skin tone, and different photos might even point to a wide range of estimated skin tone. Illumination assessment addresses this issue by guiding the user toward taking a photo under optimal illumination. In Section~\ref{sec:usecase_exp}, we will show how the shade recommendation improves with our illumination assessment.
\subsubsection{Color Comparison Using CIEDE2000 distance}
To compare the RGB value representation of a foundation shade with estimated skin tone, \emph{CIEDE2000 color difference} was used \cite{ciede2000}. This number assigns a distance between two colors in the scale of $0$ to $100$ and was developed to capture the perceptual color difference. CIEDE2000 had been tested via studies involving human observers \cite{sharma2005ciede2000, Liu:13}. Heuristically, it is accepted that the distance less than $2$ means two colors are very close to each other, and between $2$ and $5$ means the colors are similar, while greater than $10$ means the colors start to be quite different to human eyes. We will use this metric to evaluate the color variance among the estimated skin tone from the set of all images and compare that with the variance among the images that the model classifies as good illuminated. Since the foundation shade range (brown hue) tends to span a small color space, the increase in variation results in recommendation that is not specific and may even be worse than a random guess. The details will be given in Section~\ref{sec:usecase_exp}.
\vspace{-5pt}
\section{Experiments}\label{sec:experiment}
In this section, we first analyze the performance of the face illumination assessment model. Then, in Section~\ref{sec:usecase_exp}, we show how it improves the accuracy of beauty product recommendation. Section~\ref{sec:customer} is dedicated to investigate impact of this study on customers.
\subsection{Face illumination assessment}\label{subsec:assessment}
To train the proposed illumination assessment model, a synthetic dataset is constructed by transferring four skin tones (two light and two dark) to an in-house dataset of 1000 faces, followed by applying 200 illumination patterns to all the obtained face images. Therefore, the dataset consists of $1000\times 4 \times 200$ images, whose labels come from the labels defined for the 200 light patterns. Consequently 20\% of the constructed dataset is set aside for validation.

We have trained our proposed model using Pytorch~\cite{b18}, with 20 epochs, batch size of 64, and Cross-entropy loss function, where more penalty is given to the minority class which is the good illuminated case. MobileNet-v2 is pre-trained on ImageNet~\cite{b19}, and all the layers are fine-tuned using the constructed dataset, as we observed a degraded performance when fine-tuning the classification layer only. 

To test the proposed framework, we collected 674 face images at neutral lighting and bad lighting (where there is a visible shadow on the face). Having a dataset of real faces as the test set is of high importance as the model itself is trained on synthetic data, and it is crucial to validate its performance on real data.

In order to show the necessity of having illumination patterns defined on both light and dark skin people, we further designed another experiment. To this end, we asked a light skin tone subject to take photos at different locations (such as living room, office, and inside a car), light sources (lamp and window), weathers (cloudy and sunny), patterns (top, right, left, front, behind, and down), and times (morning, noon, and evening), resulted in 130 light patterns. These patterns are then transferred to the 1000 in-house faces, to train a MobileNet-v2 model. Results of this model (referred to as light skin model) is shown in table~\ref{tab:result}, along with the results of the proposed model. Note that the higher specificity of the light skin model is due to the fact that this model classifies most of the dark skin tones as bad illuminated, which in turn has resulted in a low sensitivity. 

As shown in table~\ref{tab:result}, in the second experiment, we compare the proposed framework with Reference~\cite{b1}. This Reference has trained a ResNet-50~\cite{b20} using its synthetic dataset. As it can be observed from this table, the proposed framework outperforms the other two models. More importantly, while the two other models completely fail on dark skin tones (classify most of them as bad-illuminated), ours is able to classify illumination on all skin tones.

\begin{table}
  \caption{Results of the illumination assessment experiment.}
  \vspace{-12pt}
\label{tab:result}
  \centering
  \footnotesize
  \begin{tabular}{llll}
    \toprule
        \cmidrule(r){2-4}
    &    & Metric  &   \\
    \cmidrule(r){2-4}
    Model & Accuracy     & Sensitivity  & Specificity    \\
    \midrule
    Proposed model & \textbf{79.4\%}  & \textbf{86.5\%}   & $72\%$         \\
    Reference~\cite{b1} & $58.5\%$     & $18.1\%$         &  \textbf{100\%}         \\
    Light skin model  & $69.4\%$     &  $60\%$         &   $78.9\%$  \\
    \bottomrule
  \end{tabular}
  \vspace{-8pt}
\end{table}

\subsection{Color Analysis for Beauty Recommendation}\label{sec:usecase_exp}
 \begin{table}
    \centering
    \footnotesize
  \caption{Comparison of the color differences among the estimated skin tones. Only using the well-illuminated photos significantly lowers the color difference average.}
\vspace{-8pt}
\begin{tabular}{cccc}
\vspace{-2pt}
\multirow{2}{*}{Model \#} & \multicolumn{3}{c}{Color Difference Average ($\pm$ Standard Deviation)}\tabularnewline
 & Labeled ``Good''  & Labeled ``Bad''  & All Photos\tabularnewline
\hline 
1 & 8.34$\pm$4.4 & 17.93$\pm$4.8 & 16.63$\pm$5.7\tabularnewline
2 & 4.03$\pm$2.1 & 8.63$\pm$3.0 & 8.01$\pm$3.3\tabularnewline
3 & 7.03$\pm$4.4 & 18.09$\pm$8.9 & 16.60$\pm$9.2 \tabularnewline
4 & 7.17$\pm$4.5 & 18.41$\pm$8.5 & 16.89$\pm$8.9\tabularnewline
5 & 8.44$\pm$5.0 & 17.39$\pm$6.7 & 16.19$\pm$7.2\tabularnewline
6 & 8.11$\pm$4.9 & 16.90$\pm$6.4 & 15.72$\pm$6.9\tabularnewline
\hline 
\end{tabular}
\vspace{-11pt}
\label{tab:variance}
\end{table}

\begin{table*}
    \centering
    \caption{\small{Number of Shades Recommended by Color Difference Threshold. Three products with three distinct shade range distributions are used to recommend shades using the closest match criteria, i.e. recommend the shades with the smallest distance to the skin tone. Note that the number of shades being recommended can change as well as the type of shades being recommended.}}
\vspace{-8pt}
\scalebox{0.8}{
\begin{tabular}{cccccccccc}
\multirow{3}{*}{Model \#} & Total \# of  & \multicolumn{6}{c}{Number of Shades Within the Distance Threshhold} & \multicolumn{2}{c}{\# of Overlapping Shades }\tabularnewline
 & Avaialbe Shades & \multicolumn{2}{c}{(Labeled as ``Good'' )} & \multicolumn{2}{c}{(Labeled as ``Bad'' )} & \multicolumn{2}{c}{(All Photos)} & \multicolumn{2}{c}{``Good'' vs ``Bad'' Photos}\tabularnewline
 & in Product (A,B,C) & dist.$<2$ & dist.$<5$ & dist.$<2$ & dist.$<5$ & dist.$<2$ & dist.$<5$ & dist.$<2$ & dist.$<5$\tabularnewline
\hline 
1 & \multirow{6}{*}{(39,12,17)} & (2,0,0) & (6,7,0) & (2,6,0) & (4,10,0) & (2,6,0) & (7,10,0) & (2,0,0) & (3,7,0)\tabularnewline
2 &  & (0,0,0) & (1,0,0) & (0,0,0) & (0,0,0) & (0,0,0) & (1,0,0) & (0,0,0) & (0,0,0)\tabularnewline
3 &  & (0,0,0) & (3,0,0) & (1,1,0) & (4,7,0) & (1,1,0) & (6,7,0) & (0,0,0) & (1,0,0)\tabularnewline
4 &  & (1,0,0) & (12,0,0) & (3,6,0) & (10,11,0) & (3,6,0) & (17,11,0) & (1,0,0) & (5,0,0)\tabularnewline
5 &  & (1,0,0) & (12,5,0) & (2,5,0) & (10,10,1) & (2,5,0) & (13,10,1) & (1,0,0) & (9,5,0)\tabularnewline
6 &  & (2,0,0) & (13,2,0) & (2,2,0) & (10,8,2) & (3,2,0) & (15,8,2) & (1,0,0) & (8,2,0)\tabularnewline
\hline 
\end{tabular}}
    \label{tab:rec}
\end{table*}
In this section, we present two different color analyses to access significance of using well-illuminated photos.  
\begin{figure*}
    \centering
    \includegraphics[scale = 0.055]{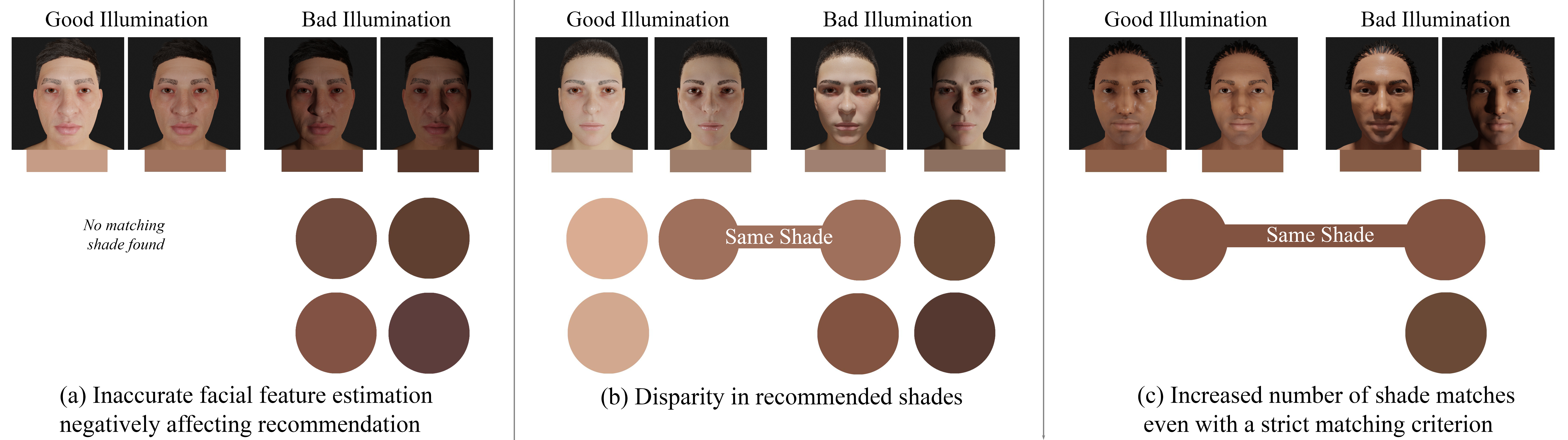}
\vspace{-10pt}
    \caption{\small{Impact of illumination conditions on shade recommendation. (a) Under the good illumination condition, the model would match with light to medium shades but this product has no offering. However, underexposed photos results in matching dark to deep shades, and the model ends up matching with nearly all shades offered in this product. (b) Only one recommended shades overlap between the well and ill-illuminated photos. Shadows can lower color saturation which may lead to such conflicting recommendation. (c) The estimated skin tones have less variance when only well-illuminated photos are used. This lead to exactly one match with the distance threshold of 2. However, if ill-illuminated photos are used, then using a low threshold does not necessarily lead to a more specific recommendation. Here, both warmer and cooler brown shades are recommended.}}
    \label{fig:dive_deep}
    \vspace{-8pt}
\end{figure*}
\subsubsection{Variation among Estimated Skin Tones}
We computed estimated skin tone for 6 different synthetically generated models representing different parts of Monk Skin Tone scale \cite{Monk_2019}. Then the distance between the estimated skin tone of the best illuminated photo and the rest was computed. This computation is to measure how much the colors deviates from the color extracted from the ideal illumination condition. Table \ref{tab:variance} summarizes the finding. The average distance is smaller if we only use the set of well-illuminated photos for all 6 models. The average color difference was greater than 10 for 5 out of 6 models if we only use the ill-illuminated images, which implies that the estimated skin tone from these have perceptually noticeable difference to the color estimated from the best illuminated photo. The examples in Figure \ref{fig:rec_big_picture} and \ref{fig:foundation}(c) verifies this: A same model can have significantly different estimated skin tones. The difference being smaller for the well-illuminated photos show that there are many ways for the illumination condition to be bad, while good illumination tend to introduce similar color patterns. 

\subsubsection{Shade Recommendation Using Face Photos}
In this experiment, we compute the distance between the shade and the estimated skin tone to simulate shade recommendation. The shade range varies by the product, and there is no industry standard on the shade range or the number of shades being offered. Therefore, we selected three products that represent different shade distributions. Product A has 39 shades of varying lightness with many medium range shades, Product B focuses on 12 deeper shades, and Product C focuses on 17 light to medium shades. To simulate the shade recommendation, we assumed that the user preference is to find the shade that is the closest to their skin tone, i.e. the color distance is minimized. Table \ref{tab:rec} shows the number of shades within the distance threshold of $2$ or $5$. We observed that using ill-illuminated photos impacts the shade recommendation in number of ways. First, the number of shades being recommended can drastically change. For Model 4, there is no shade within distance 5 for Product B if we only use the good images. However, 11 out of 12 shades are less than distance 5 if we use the bad illuminated images for the feature estimation. Recommending 11 out of 12 shades would not provide any discriminative power, and may not be any better than randomly recommending one. In addition, if the goal is to recommend a product that has the model's shade, ill-illuminated photos may recommend products that are not intended for the the model's skin tone (Figure \ref{fig:dive_deep}(a)). Next, there is a disagreement between the shade recommended by the well vs. ill-illuminated images. For Model 3, good images generate 3 distance $<5$ matches for Product A while the bad images generate 4 such. However, only 1 shade overlaps between these, meaning the user may receive a conflicting recommendation depending on the illumination condition (Figure \ref{fig:dive_deep}(b)). This implies that in the absence of illumination assessment - the input face photo can be either well or ill-illuminated - the variability in shade recommendation will be widened. This is reflected in the larger numbers for "All Photos" column in Table \ref{tab:rec}. It is also important to note that the product images tend to be taken under well-illuminated environment. Therefore, comparing the product color with the skin tone estimated from an ill-illuminated photo may lead to an incorrect shade recommendation altogether. Lastly, illumination condition impacts the number of distance $<2$ matches. In Table \ref{tab:rec}, we see there are more number of matches if we use bad illuminated images. This is another reflection of different color variation patterns introduced by varying light conditions, while color variation is more confined for the good illuminated images (Figure \ref{fig:dive_deep}(c)). Note that this may not be true if we relax the distance threshold to 5. That is, the ill-illuminated photos do not always increase the shade matches and may even decrease it as the threshold increases. This is due to over or under exposed pictures producing very light or dark browns and the shadows can lower the color saturation. A typical foundation shade ranges may not cover these ranges. 
\vspace{-9pt}
\subsection{Customer Problem Statement}\label{sec:customer}
The beauty e-commerce industry is witnessing a growing trend of applications that provide customers with personalized recommendations based on their face features. Given the diversity in mobile devices and customer skin tones and the variability of light conditions, the models using the photos may behave unexpectedly. The models are often trained under the assumption that the input data is a faithful representation of the customer's face and does not factor possible ill-illuminated environments.
Reduced performance of the model that is a part of a recommendation framework can decrease customer satisfaction. A user guidance module can solve this problem by encouraging the users to take photos under a good lighting environment. In this work, we propose a framework that can detect the quality of the illumination in face photos. We ensure the good performance of the model across various skin tones which is often neglected in previous studies. 
In our experiments, we have found that our proposed framework has a higher probability of producing successful outcomes compared to its counterparts. This indicates that the application using the face images will be more satisfactory if they leverage our approach. For example, our approach may be used to reject poorly illuminated input photos, so the customers can continue taking photos until a good illumination is detected. The customer may also upload a photo, and our model informs them of any possible issue of bad illumination. The customer then may decide if they still want to proceed or upload a new one. 
\section{Conclusion}\label{sec:conclusion}
In this work, we approached fair and responsible beauty product recommendation by proposing a light assessment model that classifies input face images as well or ill-illuminated. The motivation behind developing such model and tool is beauty product recommendations often requiring the input face photo to have good lighting. The pivotal piece in enhancing the model was to develop synthetic dataset with a wide range of skin tones and illumination patterns. This dataset synthesis also meant only 1,000 real face images were needed for our method, which provides the benefit of minimizing the manual data collection and labeling effort which can be expensive and time consuming. The synthetic dataset was used to train a MobileNet-v2 for the final illumination condition classification. Our experiments evaluated on a real dataset show that the proposed approach outperforms its counterparts. It also improves the quality of the shade recommendation and eliminates the need to correct colors which can be have biased performance and the result unexplainable to the user. The live feedback and the interactivity of our tool may provide more trustworthy experience for the users. 
\bibliographystyle{ACM-Reference-Format}
\bibliography{egbib}

\end{document}